\documentclass[11pt]{article}
\pdfoutput=1
\usepackage{acl2016}
\usepackage{multirow}
\usepackage{graphicx}
\usepackage{fancybox}
\usepackage{adjustbox}
\aclfinalcopy 

\title{Chinese Event Extraction Using Deep Neural Network with Word Embedding}

\author{Yandi Xia and Yang Liu  \\
The University of Texas at Dallas\\
 Computer Science Department \\
 {\tt {yandixia,yangl@hlt.utdallas.edu}}\\}

\date{}

\begin{document}
\maketitle

\begin{abstract}
A lot of prior work on event extraction has exploited a variety of features to represent events.
Such methods have several drawbacks: 1) the features are often specific for a particular domain and do not generalize well;
 2) the features are derived from various linguistic analyses and are error-prone; and 3) some features may be expensive
and require domain expert. In this paper, we develop a Chinese event extraction system that uses word embedding vectors to represent language, and deep neural networks to learn the abstract feature representation  in order to greatly reduce the effort of feature engineering.
In addition, in this framework, we leverage large amount of unlabeled data, which can address the problem of limited labeled corpus for this task.
Our experiments show that our proposed method performs better compared to the system using rich language features,
and using unlabeled data benefits the word embeddings.
This study suggests the potential of DNN and word embedding for the event extraction task.
\end{abstract}

\section{Introduction}
Event extraction is a task of information extraction. It is a complicated task including many sub-tasks.
In ACE, event extraction includes four sub-tasks \cite{ahn2006stages,grishman2005nyu}: event trigger identification, trigger type classification, argument identification, and argument role classification.
In this paper, we focus on trigger identification on Chinese.
Event trigger identification  is often the first step of an event extraction system.
It identifies the words that most explicitly indicate the occurrence of events. For example, in this sentence: Yahoo announced its acquisition of KIMO website. Here `acquisition' is an event trigger, and it triggers one of the sub-types of BUSINESS event type -- Merge-Org event.

There are some trends in the research for event extraction in the past years.
First, much prior work has focused on exploiting rich language features like syntax, part-of-speech (POS), parsing structure,
named entities, synonym dictionaries, etc.
For example, \newcite{chen2009language} and \newcite{chen2012joint} both used a variety of such features, achieving the state-of-the-art performance for Chinese event extraction. In English corpus, the earliest work on ACE corpus \cite{ahn2006stages} also designed a lot of features for this task and set the baseline of English event extraction.
However, these systems have some common drawbacks:
a) those features vary from corpus to corpus, and language to language. One almost needs to setup the whole system for new data. For example, \cite{chen2009language} designed a character based system with character based features for Chinese event extraction, in order to reduce the errors caused by word segmentation. However, such features make no sense for English corpora. On the other hand, in the English systems, features are needed to represent phenomena such as tenses, whereas in Chinese, this is not a problem.
Same thing happens if one wants to migrate an event extraction system from news article domain to biomedical domains.
Most of the features designed previously are not very useful for the new data.
b) those features depend on some fundamental NLP modules (e.g., POS tagging, parsing, named entity recognition),
which are not perfect and introduce a lot of noise in the derived features.
This is especially true for Chinese (and languages with low resources), in which many NLP tools do not perform well.
c) some of features are expensive to acquire, because expert knowledge or resources are required, for example,
synonym dictionaries.

Second, due to the lack of labeled data for event extraction, more and more research started focusing on semi-supervised learning \cite{liao2010filtered,ali2014event,liemploying},
unsupervised learning \cite{rusu2014unsupervised}, and distantly supervised learning \cite{reschkeevent}.
These studies show that additional unlabeled data is very useful to acquire more information or understand about language.
For example, large unlabeled data can be used to learn the patterns for  event extraction  \cite{liao2010filtered,huang2012bootstrapped}.

Motivated by the need to overcome the problem with designing features and the potential benefit of unlabeled data, in this pilot study, we set to answer three questions:
(1) can we get around feature engineering in event trigger detection using the deep neural networks and word embedding?
(2) is word embedding a better representation for this task than the large set of carefully crafted discrete features?
(3) can we effectively leverage unlabeled data for word embedding in the trigger detection task?

Word embedding has been proved to be very successful combining with deep learning, the increasingly popular learning method.
After \newcite{collobert2011natural} and \newcite{socher2012deep} brought up a unified deep structure for NLP tasks, much work using this combination has emerged to challenge the traditional feature based methods.
\newcite{collobert2011natural} showed great performance in tasks such as part-of-speech-tagging, chunking, named entity recognition, and semantic role labeling with one unified deep structure, which is comparable to those feature based methods. \newcite{li-liu:2014}'s work applied word embedding on text normalization task which use the similarity between word vector to represent the sematic relationship between two words.
\newcite{qi2014deep} and \newcite{zheng2013deep} adopted this structure for Chinese NLP tasks, and beat the state-of-the-art performance in word segmentation, part-of-speech-tagging and named entity recognition tasks.
In addition, \newcite{li2015using} applied word embedding on two large corpus, one is a set of news articles and the other is their corresponding summaries. Then for each token in two word embedding, they design additional features to help estimate their importance for final summary. The experiments showed that the
feature based on word embedding are very useful.
Inspired by these successful efforts, in this work we design a deep structure event extraction system which takes word embedding representation of Chinese data.
We expect that deep learning can learn abstract feature representation and
word embedding can effectively represent semantic and syntactic similarity between words, and thus
can help identify new event trigger words that are not in the training examples.
For example,
if beat is a trigger word in the training set, it is hard to use synonym information to determine word attack is a trigger in the test data.
However, word embedding may be able to find such semantic similar words, in either a supervised or unsupervised fashion,
 and improve system's recall.
To our knowledge, there is no prior work that has explored the use of word embedding and deep learning for Chinese event extraction.

In this work, we build a deep neural network model which represents Chinese characters with word embedding vectors.
To leverage unlabeled data, we generate word embeddings from that and use them for
pre-training in DNN.
We evaluate our methods on the Chinese ACE corpus.
Different from previous work that used ground truth information such as named entities,
time and value labels,
we use a more realistic setup with such information automatically generated.
Our results show that we can achieve
better performance using DNN than a feature-based maximum entropy classifier
for event trigger detection, and then using unlabeled data for word embedding pretraining
has additional benefit.

The rest of the paper is structured as follows.
Section 2 describes the event trigger identification methods, including the maximum entropy classifier that uses
a set of rich features, and our DNN and word embedding system.
Section 3 shows the experimental results.
Conclusions and future work appear in Section 4.

\section{Event Trigger Identification}

The follow introduces the two methods we use for event trigger word identification.

\subsection{Baseline}

Following the works of \cite{chen2009language,chen2012joint}, we build the feature based baseline system.
As observed in \cite{chen2009language}, there is a considerable portion of segmentation errors because of the low performance of the Chinese word segmentation tool and the ambiguity in human labels.
Therefore, we model it as a
sequence-labeling task and use the BIO encoding
method (each character either begins, is inside, or outside
of a trigger word)
As for the features, both papers did a lot in feature analysis, therefore we adopt most of their features, listed below:
\begin{itemize}
\itemsep=0.01cm
\item \textbf{Lexical features:} current character; the characters that surround it; current POS; and the POS that surround it; the combination of current character and its POS.
\item \textbf{Syntactic features:} depth of the current character in the parse tree; path from the current node to the root; the phrase structure expanded from the parent node; phrase type.
\item \textbf{Semantic dictionary:} whether it is in the predicate list from \cite{xue2009adding}; synonym entry number of the current character.
\item \textbf{Nearest entity:} the type of the left nearest entity to the current character; the type of the right nearest entity to the current character. Here, distance is measured by text length.
\end{itemize}
In \cite{chen2009language,chen2012joint}, human labeled named entity information was used.
In this study, we use a more realistic setup --  we use automatically identified named entities in our features.
In our experiments, we will evaluate the effect of using such imperfect features.

\subsection{DNN Model}

\begin{figure}[ht]
\begin{center}
\includegraphics[width=2.8in]{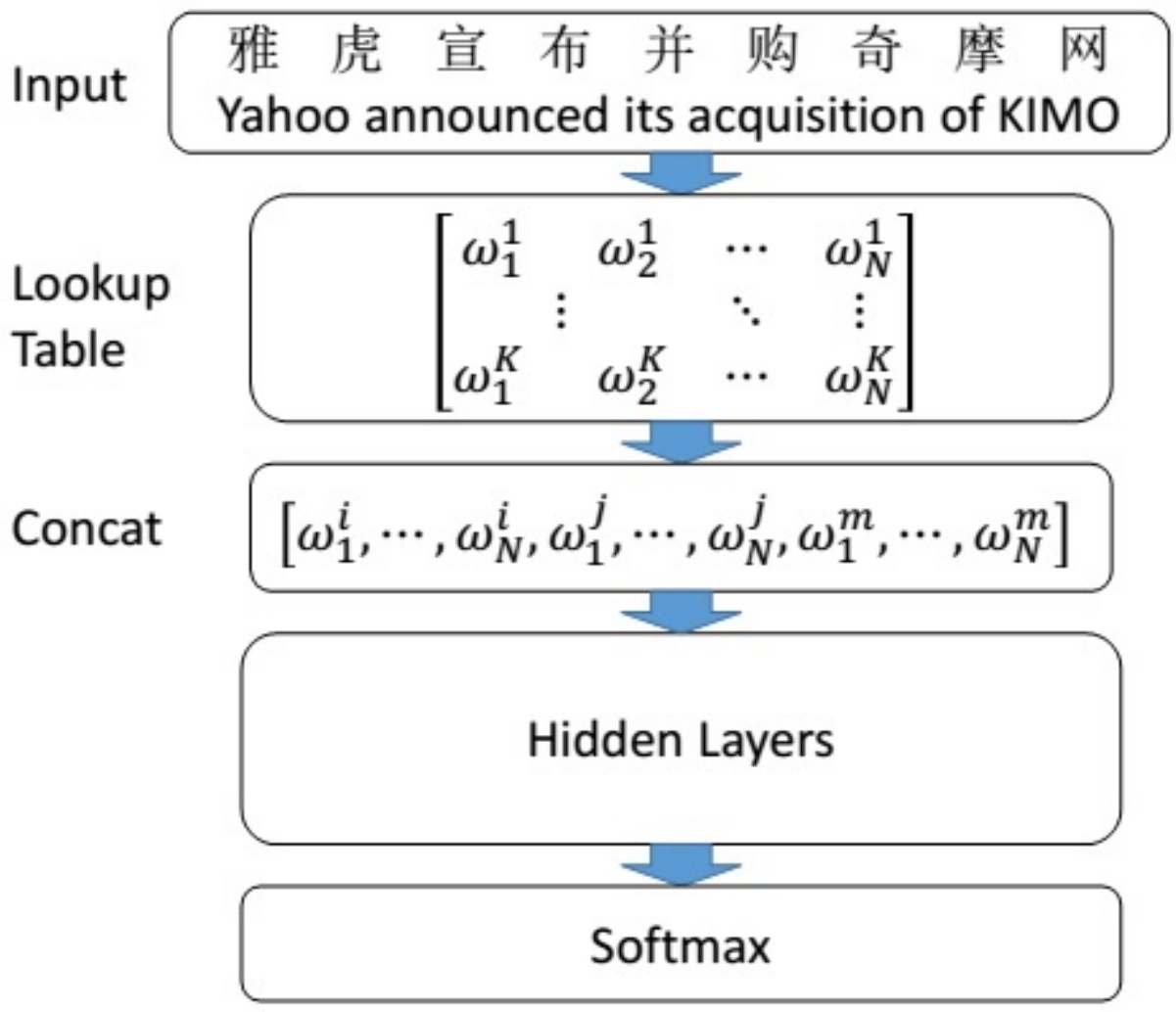}
\caption{ Structure of DNN model. }
\label{fig:dnnstr}
\end{center}
\vspace{-0.1in}
\end{figure}

We followed the work of \cite{collobert2011natural} and designed a similar DNN architecture, as shown in Figure \ref{fig:dnnstr}. Considering the word segmentation problem in Chinese, we also choose the window based character stream as input.
Here we model the input character as its index in the character dictionary.
The first layer is a lookup table whose entries are word embedding vectors. Then we concatenate those selected vectors and pass it to the hidden layers.
In the hidden layers we use non-linear Tanh as the activation function.
At the top of the structure, we put softmax  to output the probabilities of the character being a part of a trigger.
Note in this method, we do not use any linguistically motivated features. The input used in this DNN is just word embedding vectors.

All the weights in the DNN including the word embedding vectors are trained automatically in the back-propagation process using the gradient descent algorithm.
During testing, the DNN system gives the probabilities of each character as BIO tag, we
treat them as emission probabilities, together
with the transition probability from baseline CRF system, we use Viterbi
decoding process to conduct the final prediction. \newcite{liimproving} and \newcite{liijcai} showed that using separated but well trained weight in Viterbi decoding can give improvement in certain conditions.
The DNN tool we use is python based software Theano \cite{Bastien-Theano-2012}.

\subsection{Using Unlabeled Data}
Unlabeled data contains a lot of abstract syntax and semantic information that can be very useful to NLP tasks.
In order to take advantage of unlabeled data (simply Chinese sentences, no event labels), we first use the RNN based word2vec toolkit from to generate the initial word vector dictionary as pre-training.
These vectors then will act like part of the DNN weights and change their values in the supervised learning progress
via back propagation.

\section{Experiments}
\noindent

The corpus we use is ACE 2005 Chinese corpus. There are 633 documents in the corpus.
We randomly choose 66 documents as the test data, and 567 documents as training data, which is similar to \cite{chen2009language}.
For performance metric, we use precision, recall, and F1 score.
If the offset and word length of the identified character chunk exactly match the gold value,
we consider that the corresponding trigger is correctly identified.
The unlabeled data we use is from \cite{graff2005chinese}.
It contains 2,466,840 Chinese newswire documents, totaling 3.9 GB. We use 100K documents of them.

Table \ref{table:result} shows the trigger identification results for different methods.
For the baseline system using ME classifier, we show two results.
One is using the named entities obtained from the Stanford coreNLP tool,
and the other one uses ground truth NER labels.
It is clear that ground truth NER information can boost the performance considerably. However, such accurate information is very hard to get in real world cases.

For the DNN model, we also report two results,  without using the unlabeled data to pretrain the network,
vs. the one that takes advantage of large amount of unlabeled data for pretraining.
We can see that pretraining improves the system performance significantly.
In addition, during the experiments, we noticed that DNN without pretraining converges far more slower than that with pretraining.

Comparing the results using the DNN and the ME classifier, we can see even without using the unlabeled data,
the DNN results are better than the feature-based ME classifier (using automatically generated NE information).
This suggests that no feature engineering is required in the DNN model -- it simply uses the character embedding
to generate more effective abstract feature representation for this classification task.
When unlabeled data is incorporated, the DNN performance is much better, even outperforming using the
reference NE information in the ME classifier.
In addition, the improved DNN results over the ME method are because of the higher recall,
 which is  consistent with our expectation -- using word embedding vectors can find semantically similar words.

\begin{table}[ht]
\begin{center}
\begin{adjustbox}{max width=8.2cm}
\begin{tabular}{c|c|c|c} \hline
  & recall & precision & F1-score \\ \hline
CRF baseline & \multirow {2}{*}{48.09} & \multirow {2}{*}{71.37} & \multirow {2}{*}{57.47} \\
auto NE &&  & \\ \hline
CRF baseline & \multirow {2}{*}{52.17} & \multirow {2}{*}{74.71} & \multirow {2}{*}{61.44} \\
ref NE  &&&  \\ \hline
DNN  & 51.36 & 66.78 & 58.06 \\ \hline
DNN  & \multirow {2}{*}{61.14} & \multirow {2}{*}{67.77} & \multirow {2}{*}{64.29} \\
with unlabeled data &&&  \\ \hline \hline
\end{tabular}
\end{adjustbox}
\end{center}
\caption{Event trigger word identification results using different systems.}
\label{table:result}
\vspace{-0.3cm}
\end{table}

\begin{figure}[ht]
\begin{center}
\includegraphics[width=2.8in]{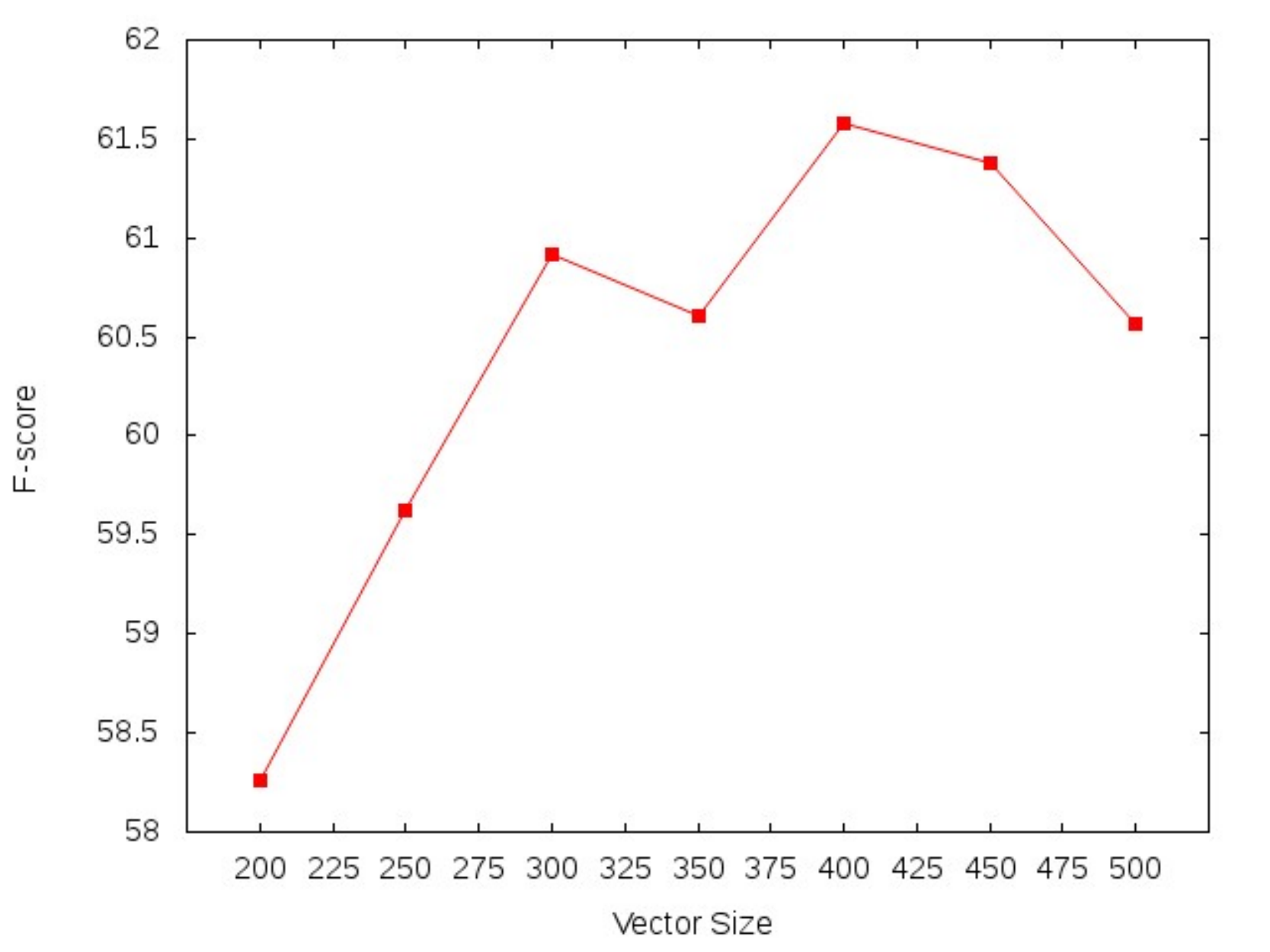}
\caption{Trigger word identification results using DNN with different character vector sizes.}
\label{fig:size}
\end{center}
\vspace{-0.1in}
\end{figure}

Figure \ref{fig:size} shows the F1 score when varying the vector size of the character embeddings.
This pattern is similar to that for other tasks. When the size is too small, it cannot represent detailed information of the
language characteristics; when it is too big, the system has too many parameters and loses the power of the
abstract representation.

\section{Conclusions and Future Work}
\noindent

In this paper, we applied word embeddings  to represent Chinese characters.
Our result on the event trigger identification task  shows that it is a better representation compared to human designed language specific features. We also show that the combination of word embedding and DNN outperform the classifier that relies on
a large set of linguistic features, and that this framework can effectively leverage unlabeled data to improve system performance.
This is the first study exploring deep learning and word embedding for Chinese event extraction.
In our current work, we use a relatively small window of characters as  the input of the DNN.
For future work, we plan to find a way to model longer context in DNN for event extraction. Furthermore,
we will move on CNN and RNN architecture for this task.

\section*{Acknowledgments}

\bibliography{acl2016}
\bibliographystyle{acl2016}

\end{document}